\documentclass[11pt]{article}
\usepackage{coling2020}

\usepackage{times}
\usepackage{url}
\usepackage{latexsym}

\usepackage{mathptmx}
\usepackage{graphicx}
\usepackage{tabularx,ragged2e}
\usepackage{tabulary}
\usepackage{adjustbox}
\usepackage{amsmath}
\usepackage{multirow}
\usepackage{floatrow}
\usepackage{url}
\usepackage{hhline}
\usepackage{multicol}
\usepackage{booktabs}
\usepackage{caption}
\usepackage{subcaption}
\usepackage[hidelinks]{hyperref}
\usepackage{fancyhdr}
\usepackage{amsmath}
\usepackage{amssymb}
\usepackage{xspace}
\usepackage{color, colortbl}

\definecolor{Gray}{gray}{0.945} %

\newcommand{\methodsCountText}{seven~} %

\newcommand{\relCount}{ten\xspace}

\newcommand{\acl}{ACL Anthology\xspace}
\newcommand{\cord}{CORD-19\xspace}
\newcommand{\totalSamplesCount}{172,073\xspace} %

\colingfinalcopy %

\title{Aspect-based Document Similarity for Research Papers}

\author{
Malte Ostendorff$^{1,2}$ \quad Terry Ruas$^{3}$  \quad Till Blume$^{4}$  \quad Bela Gipp$^{2,3}$  \quad Georg Rehm$^{1}$ \\
$^{1}$German Research Center for Artificial Intelligence, Berlin, Germany \\
$^{2}$University of Konstanz, Konstanz, Germany  \\
$^{3}$University of Wuppertal, Wuppertal, Germany \\
$^{4}$University of Kiel, Kiel, Germany \\
{\tt malte.ostendorff@dfki.de} \\
}

\date{}

\begin{document}
\maketitle
\begin{abstract}

Traditional document similarity measures provide a coarse-grained distinction between similar and dissimilar documents.
Typically, they do not consider in what aspects two documents are similar. 
This limits the granularity of applications like recommender systems that rely on document similarity.
In this paper, we extend similarity with aspect information by performing a pairwise document classification task.
We evaluate our aspect-based document similarity for research papers.
Paper citations indicate the aspect-based similarity, i.e., the section title in which a citation occurs acts as a label for the pair of citing and cited paper.
We apply a series of Transformer models such as RoBERTa, ELECTRA, XLNet, and BERT variations and compare them to an LSTM baseline.
We perform our experiments on two newly constructed datasets of \totalSamplesCount research paper pairs from the \acl and \cord corpus. 
Our results show SciBERT as the best performing system. 
A qualitative examination validates our quantitative results.
Our findings motivate future research of aspect-based document similarity and the development of a recommender system based on the evaluated techniques.
We make our datasets, code, and trained models publicly available.

\end{abstract}

\section{Introduction}
\label{sect:intro}
\blfootnote{
    \hspace{-0.65cm}  %
     This work is licensed under a Creative Commons 
     Attribution 4.0 International License.
     License details:
     \url{http://creativecommons.org/licenses/by/4.0/}.
    
}

\begin{figure}[ht]
\centering
\begin{subfigure}{.49\textwidth}
\centering
\includegraphics[page=3,clip,width=0.85\linewidth,trim={2cm 2.3cm 2cm 1.15cm},clip]{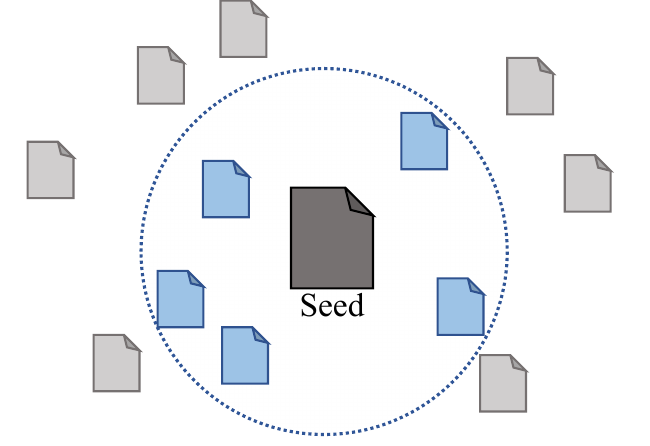}
\caption{\label{fig:docsim} Aspect-free similarity.}
\end{subfigure}
\begin{subfigure}{.49\textwidth}
\centering
\includegraphics[page=5,clip,width=0.85\linewidth,trim={2cm 2.3cm 2cm 1.15cm},clip]{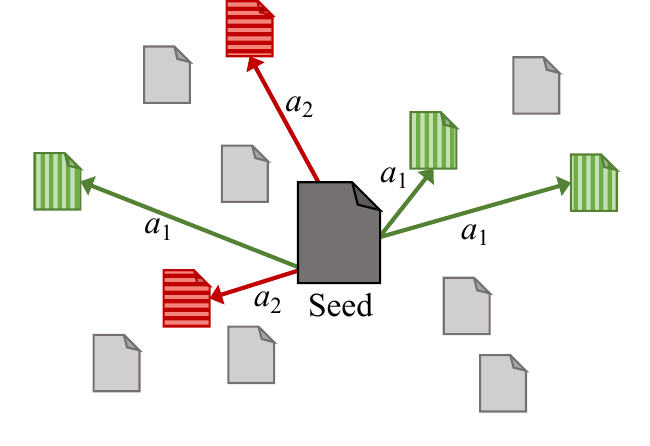}
\caption{\label{fig:docrel} Aspect-based similarity.}
\end{subfigure}

\caption{\label{fig:docsim-vs-docrel}Most RS rely on similarity measures between a seed and the $k$ most similar target documents (a). This neglects the aspects which make two or more documents similar. In aspect-based document similarity (b), documents are related according to the inner aspects connecting them ($a_1$ or $a_2$).}

\end{figure}

Recommender systems (RS) assist researchers in finding relevant papers for their work.
When user feedback is sparse or unavailable, content-based approaches and corresponding document similarity measures are employed~\cite{Beel2016}.
RSs recommend a candidate document depending on whether it is similar or dissimilar to the seed document. 
This coarse-grained similarity assessment (similar or not) neglects the many facets that can make two documents similar. 
Concerning the general concept of similarity, \newcite{Goodman1972}, and \newcite{Bar2011} even argue that similarity is an ill-defined notion unless one can say to what aspects the similarity relates.
In RS for scientific papers, the similarity is often concerned with multiple facets of the presented research, e.g., method, findings~\cite{Huang2020}.
Given the document similarity can differentiate research aspects, one could obtain specific tailored recommendations. 
For instance, a paper with similar methods but different findings could be recommended. 
Such a RS would facilitate the discovery of analogies in research literature~\cite{Chan2018}.
We describe the underlying multiple aspect similarity in research papers as \textit{aspect-based document similarity}.
Figure~\ref{fig:docsim-vs-docrel} illustrates the aspect-based in contrast to aspect-free similarity (traditional).
Following the research paper example, aspect $a_1$ concerns findings and aspect $a_2$ methods (red and green in Figure~\ref{fig:docrel}). 

In prior work \cite{Ostendorff2020}, we propose to infer an aspect for document similarity formulating the problem as a multi-class classification of document pairs.
We extend our prior work to a multi-label scenario and focus on scientific literature instead of a general one (Wikipedia articles).
Similar to the work of \newcite{Jiang2019} and \newcite{Cohan2020}, we use citations as training signals. 
Instead of using citations for binary classification (i.e., similar and dissimilar), we include the title of the section in which a citation occurs, as a label for a document pair.
The section titles of citations describe the aspect-based similarity of citing and cited papers.
Our datasets originate from ACL Anthology~\cite{Bird2008} and CORD-19~\cite{Wang2020}.

In summary, our contributions are: 
(1) We extend traditional document similarity to aspect-based in a multi-label multi-class document classification task.
(2) We demonstrate the aspect-based document similarity is well-suited for research papers.
(3) We evaluate six Transformer-based models and a baseline for the pairwise document classification task. 
(4) We publish our source code, trained models, and two datasets from the computer linguistic and biomedical domain, to foster further research.

\section{Related Word} \label{sect:related_work}

In the following, we discuss work on text similarity, recommendation, and applications of Transformers.

\newcite{Bar2011} discuss the notion of similarity as often ill-defined in the literature and used as an ``umbrella term covering quite different phenomena''. 
\newcite{Bar2011} also formalize what text similarity is and suggest content, structure, and style are the major dimensions inherent to text.
For literature recommendation, the content and user information are the most predominant dimension to consider~\cite{Beel2016}.

\newcite{Chan2018} explore aspect-based document similarity as a segmentation task instead of a classification task.
They segment the abstracts of collaborative and social computing papers into four classes, depending on their research aspects: background, purpose, mechanism, and findings. 
Cosine similarity computed on segment representations allows the retrieval of similar papers for a specific aspect.
\newcite{Huang2020} apply the same segmentation approach on the CORD-19 corpus~\cite{Wang2020}.
\newcite{Kobayashi2018} follow a related approach for citation recommendations. They classify sections into discourse facets and build document vectors for each facet. 
Nevertheless, segmentation is a suboptimal alternative as it breaks the coherence of documents. 
With pairwise document classification, the similarity is aspect-based without sacrificing the document coherence.

Our experiments investigate Transformer language models~\cite{Vaswani2017}.
BERT~\cite{Devlin2019}, RoBERTa~\cite{Liu2019}, XLNet~\cite{Yang2019}, and ELECTRA~\cite{Clark2020} improve many NLP tasks, e.g., natural language inference~\cite{Bowman2015,Williams2018} and semantic textual similarity~\cite{Cer2017}.
\newcite{Reimers2019} demonstrate how BERT models can be combined in a Siamese network~\cite{Bromley1993} to produce embeddings that can be compared using cosine similarity.
\newcite{Adhikari2019} and~\newcite{Ostendorff2019} explore BERT for the classification of single documents with respect to sentiment or topic. 
\newcite{Beltagy2019} and~\newcite{Cohan2020} study domain-specific Transformers for NLP tasks on scientific documents.

Moreover,~\newcite{Cohan2020} are the first to use Transformers to encode titles and abstracts of papers to generate recommendations.
\newcite{Hassan2019} also use BERT for RS, but only to encode paper titles.
Other recent RSs rely on other techniques such as co-citation analysis, TF-IDF, or Paragraph Vectors~\cite{Kanakia2019,Collins2019}.

In prior work~\cite{Ostendorff2020}, we model aspect-based similarity as a pairwise multi-class document classification task.
We use the edges from the Wikidata knowledge graph as aspect information for the similarity of Wikipedia articles.
The used task definition allows only a single-label classification.
For research papers, this definition is not adequate.
Two papers can be similar in multiple aspects. 
Accordingly, we incorporate the multi-class classification task and expand it to a multi-label one.

For our experiments, we utilize citations and the section titles in that the citations occur as classification labels. 
\newcite{Nanni2018} demonstrate a related approach in the context of entity linking.
They argue that in many situations a link to an entity offers only relatively coarse-grained semantic information.
To account for different aspects in that an entity is mentioned, \newcite{Nanni2018} link the entities not only to their respective Wikipedia articles but also to the sections that represent the different aspects.

With segment-level similarity and pairwise multi-class single-label classification, preliminary approaches addressing the aspect-based similarity are available.
In particular, Transformer models seem promising with their success for similarity, classification, and other related tasks.

\section{Experiments} \label{sec:experiment} %
\begin{figure}[ht]
\centering
\begin{subfigure}{.49\textwidth}
\centering
\includegraphics[page=1,clip,width=0.99\linewidth,trim={0.35cm 0.3cm 0.4cm 0.6cm},clip]{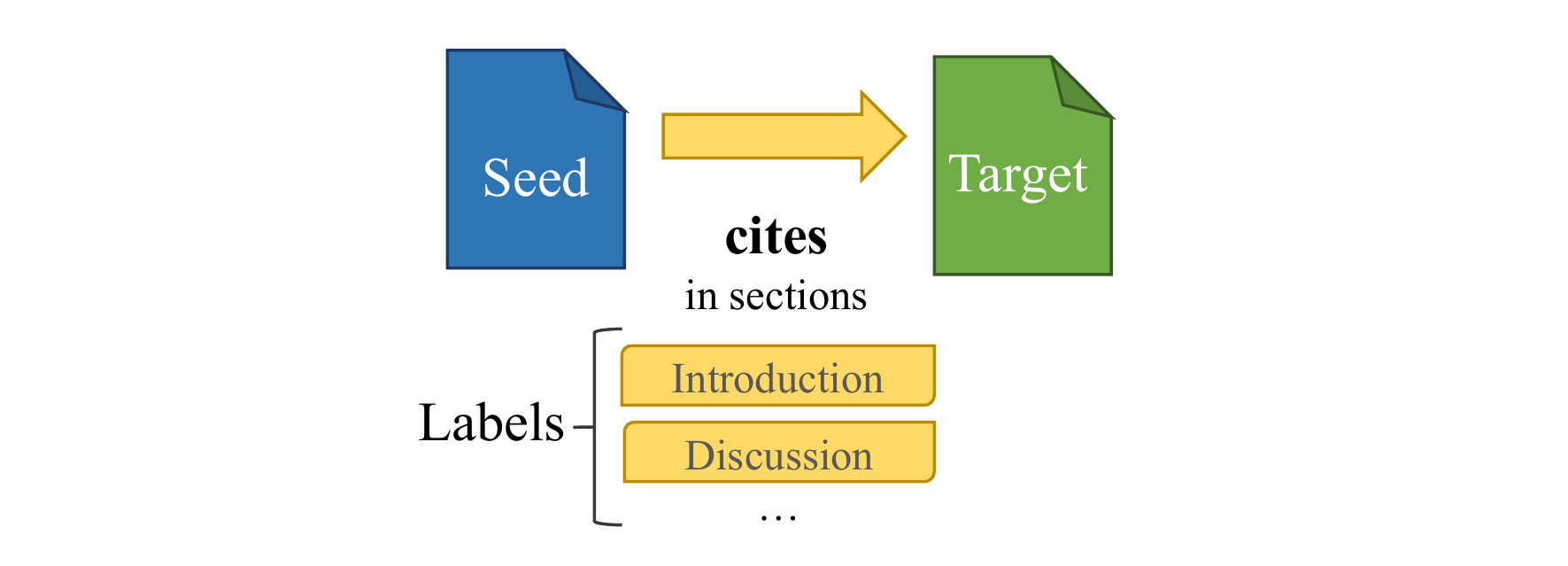}
\caption{\label{fig:dataset} Dataset.}
\end{subfigure}
\begin{subfigure}{.49\textwidth}
\centering
\includegraphics[page=2,clip,width=0.99\linewidth,trim={0.35cm 0.1cm 0.4cm 0cm},clip]{figures/dataset-and-system.pdf}
\caption{\label{fig:transformer} Document Pair Classification.}
\end{subfigure}

\caption{\label{fig:method} We use the section titles from citations as labels for document pairs. The sections define the aspects of the similarity. A Transformer model with titles and abstracts as input is used for classification.} 
 
\end{figure}

We present our methodology (Figure~\ref{fig:method}) for classifying the aspect-based similarity of research papers.

\subsection{Datasets} \label{ssec:dataset}

The generation of human-annotated data for research paper recommendations is costly and usually limited to small quantities~\cite{Beel2016}.
The small dataset size hampers the application of learning algorithms.
To mitigate the data scarcity problem, researchers rely on citations as ground truth, i.e., when a citation exists between two papers, the two papers are considered similar~\cite{Jiang2019,Cohan2020}. 
Whether a citation exists or does not correspond to a label for a binary classification.  
To make the similarity aspect-based, we transfer this idea to the problem of multi-label multi-class classification.
As ground truth, we adopt the title of the section in which the citation from paper $A$ (seed) to $B$ (target) occurs as label class (Figure~\ref{fig:dataset}). 
The classification is multi-class because of multiple section titles, and multi-label because paper A can cite B in multiple sections.
For example, paper $A$ citing $B$ in the \textit{Introduction} and \textit{Discussion} section would correspond to one sample of the dataset.

\paragraph{ACL Anthology}

We adopt the ACL Anthology Reference Corpus~\cite{Bird2008} as dataset.
The corpus comprises 22,878 research papers about computational linguistics.
Aside from full-texts, the ACL Anthology dataset provides additional citation data.
The citations are annotated with the title of the section in which the citation markers are located.
This information is required for our experiments.

\paragraph{CORD-19}

The COVID-19 Open Research Dataset (CORD-19) is a collection of papers on COVID-19 and related coronavirus research from several biomedical digital libraries~\cite{Wang2020}.
The citation and metadata of all CORD-19 papers are standardized according to the processing pipeline of~\newcite{Lo2019}.
Citations in CORD-19 are also annotated with section titles.

\subsection{Data Preprocessing} \label{ssec:preprocessing}

\begin{table}[ht]
\begin{subtable}{.49\textwidth}
\centering
\small
\begin{tabular}{lrlr}
\toprule
\textbf{Label class} & \textbf{Count} & \textbf{Label class} & \textbf{Count} \\
\midrule
Introduction         & 16,279         & Conclusion           & 1,158          \\
Related Work         & 12,600         & Discussion           & 1,132          \\
Experiment           & 4,025          & Evaluation           & 971            \\
Background           & 1,365          & Methods               & 719            \\
Results              & 1,181          & \textit{Other}       & 22,249    \\\bottomrule    
\end{tabular}
\caption{ACL Anthology}

\end{subtable}
\begin{subtable}{.49\textwidth}
\centering
\small
\begin{tabular}{lrlr}
\toprule
\textbf{Label class} & \textbf{Count} & \textbf{Label class} & \textbf{Count} \\
\midrule
Introduction         & 15,108         & Background           & 454          \\
Discussion         & 13,258         & Materials           & 420          \\
Conclusion           & 1,003          & Virus           & 218            \\
Results           & 910          & Future work               & 171            \\
Methods              & 523          & \textit{Other}       & 43,154    \\\bottomrule    
\end{tabular}
\caption{CORD-19}

\end{subtable}

\caption{\label{tab:dataset_labels}Label class distribution as extracted from citing section titles in the two datasets. We report the top nine section-classes in decreasing order, and group the remaining as \textit{Other}.}

\end{table}

Considering the \acl and \cord, we derive two datasets for pairwise multi-label multi-class document classification. 
The section titles of the citations, i.e., the label classes, are presented in Table~\ref{tab:dataset_labels}.
We normalize sections titles (lowercase, letters-only, singular to plural) and resolve combined sections into multiple ones (\textit{Conclusion and Future Work} to \textit{Conclusion}; \textit{Future Work}). 
We query the API of DBLP~\cite{Ley2009} and Semantic Scholar~\cite{Lo2019} to match citation and retrieve missing information from the papers such as abstracts. 
Invalid papers without any text or duplicated ones are also removed.
We divide both datasets, \acl and \cord, in ten classes according to their number of samples, whereby the first nine compose the most popular section titles and the tenth (\textit{Other}) groups the remaining ones.
Even though the decision of our ten classes might neglect section title variations in the literature, our model still doubles the number of research aspects \newcite{Huang2020} and \newcite{Chan2018} defined.
The resulting class distribution is unbalanced but it reflects the true nature of the corpora as Table~\ref{tab:examples} shows. %
Scripts for reproducing the datasets are available with our source code.

\subsection{Negative Sampling} \label{ssec:negative}

In addition to the \relCount positive classes (Table~\ref{tab:dataset_labels}), we introduce a class named \textit{None} that works as a negative counterpart for our positive samples in the same proportion~\cite{Mikolov2013}. 
The \textit{None} document pairs are randomly selected and are dissimilar.
A random pair of papers is a negative sample when the papers do not exist as positive pair, are not co-cited together, do not share any authors, and are not be published in the same venue.
We generate 24,275 negative samples for \acl and 33,083 for \cord.
These samples let the models distinguish between similar and dissimilar documents.

\subsection{Systems} \label{ssec:systems}

We focus on sequence pair classification with models based on the Transformer architecture~\cite{Vaswani2017}. 
Transformer-based models are often used in text similarity tasks~\cite{Jiang2019,Reimers2019}.
Moreover, \newcite{Ostendorff2020} found vanilla Transformers, e.g., BERT~\cite{Devlin2019}, XLNet~\cite{Yang2019}, outperform Siamese networks~\cite{Bromley1993} and traditional word embeddings (e.g., GloVe~\cite{Pennington2014}, Paragraph Vectors~\cite{Le2014}) in the pairwise document classification task. 
Hence, we exclude Siamese networks and pretrained word embeddings models in our experiments. 
Instead, we investigate six Transformer variations and an additional baseline for comparison.
The titles and abstracts of research paper pairs are used as input into the model, whereby the \texttt{[SEP]} token separates seed and target paper (Figure~\ref{fig:transformer}). 
This procedure is based on our prior work~\cite{Ostendorff2020}.
We do not use full-texts in our experiments as many papers are not freely available and the selected Transformer models impose a hard limit of 512 tokens.

\paragraph{Baseline LSTM}
As a baseline, we use a bidirectional LSTM~\cite{Hochreiter1997}.
To derive representations for document pairs, we feed the title and abstract of two papers through the LSTM, 
whereby the papers are separated with a special separator token.
We use the SpaCy tokenizer~\cite{spacy2} and word vectors from fastText~\cite{Bojanowski2017}. %
The word vectors are pretrained on the abstracts of ACL Anthology or CORD-19 datasets. %

\paragraph{BERT, Covid-BERT \& SciBERT}

BERT is a neural language model based on the Transformer architecture~\cite{Devlin2019}.
Commonly, BERT models are pretrained on large text corpora in unsupervised fashion.
The two pretraining objectives are the recovery of masked tokens (i.e., mask language modeling) and next sentence prediction (NSP).
After pretraining, BERT models are fine-tuned for specific tasks like sentence similarity~\cite{Reimers2019} or document classification~\cite{Ostendorff2019}. 
Several BERT models pretrained on different corpora are publicly available. %
For our experiments, we evaluate three BERT variations. 
(1) The BERT model from \newcite{Devlin2019}, trained on English Wikipedia and the BooksCorpus~\cite{Zhu2015}. 
(2) SciBERT~\cite{Beltagy2019}, a variation of BERT tailored for scientific literature, which is pretrained on computer science and biomedical research papers. 
(3) Covid-BERT~\cite{Chan2020} is the original BERT model from \newcite{Devlin2019} but fine-tuned on the \cord corpus.

BioBERT~\cite{Lee2019} is another BERT model specialized in the biomedical domain. 
Nonetheless, we exclude BioBERT of our experiments as SciBERT outperforms it on biomedical tasks~\cite{Beltagy2019}.
We also omit the BERT variations from \newcite{Cohan2020} since they use citation during pretraining risking data leakage into our test set.
All three models, i.e., BERT, SciBERT, and Covid-BERT, are similar in their structure, except for the corpus used during the language model training. %

\paragraph{RoBERTa}

\newcite{Liu2019} propose RoBERTa, which is a BERT model trained on larger batches, longer training time, and drops the NSP task from its objective. %
Moreover, RoBERTa uses additional corpora for pretraining, namely Common Crawl News~\cite{Nagel2016}, OpenWebText~\cite{Gokaslan2019OpenWeb}, and STORIES~\cite{Trinh2018}.

\paragraph{XLNet}

Unlike BERT, XLNet~\cite{Yang2019} is not an autoencoder but an autoregressive language model.
XLNet does not employ NSP. %
We use a XLNet model published by its authors, which is pretrained on Wikipedia, BooksCorpus~\cite{Zhu2015}, Giga5~\cite{Parker2011}, ClueWeb 2012-B~\cite{Callan2009}, and Common Crawl~\cite{Elbaz2007}.

\paragraph{ELECTRA}

ELECTRA~\cite{Clark2020} has in addition to mask language modelling the pretraining objective of detecting replaced tokens in the input sequence. 
For this objective, \newcite{Clark2020} use a generator that replaces tokens and a discriminator network that detects the replacements.
The generator and discriminator are both Transformer models.
ELECTRA does not use the NSP objective. 
For our experiments, we use the discriminator model of ELECTRA.
The pretrained ELECTRA discriminator model is pretrained on the same data as BERT. %

\paragraph{Hyperparameters \& Implementation}

We choose the LSTM hyperparameters according to the findings of~\newcite{Reimers2017} as follows: 10 epochs for training, batch size $b=8$, learning rate $\eta=1^{-5}$, two LSTM layers with 100 hidden size, attention, and dropout with probability $d=0.1$.
While the LSTM baseline uses vanilla PyTorch, all Transformer-based techniques are implemented using the Huggingface API~\cite{Wolf2019}. 
Each Transformer model is used in its \texttt{BASE} version.
The hyperparameters for Tranformer fine-tuning are aligned with \newcite{Devlin2019}: four training epochs, learning rate $\eta=2^{-5}$, batch size $b=8$, and Adam optimizer with $\epsilon=1^{-8}$. 
We conduct the evaluation in a stratified $k$-fold cross-validation with $k=4$ (i.e., the class distribution remains identical for each fold).
This results, on average, in 54,618.75/18,206.25 train/test samples for \acl, and 74,436/24,812 train/test samples for \cord.
The source code, datasets, and trained models are publicly available\footnote{GitHub repository: \url{https://github.com/malteos/aspect-document-similarity}}.
We provide a Google Colab to try out the trained models on any papers from Semantic Scholar \footnote{\url{https://colab.research.google.com/github/malteos/aspect-document-similarity/blob/master/demo.ipynb}}.

\newpage %

\section{Results} \label{sec:results} %

Our results are divided in three parts: overall, label classes, and qualitative evaluation\footnote{The label and qualitative evaluations discard one of the two datasets due to space constraints but are available on GitHub.}.

\subsection{Overall Evaluation} \label{ssec:overall}

The overall results of our quantitative evaluation are presented in Table~\ref{tab:overall_results}.
We conduct the evaluation as 4-fold cross validation based on our datasets.
We report micro and macro average for precision, recall, and F1-score to account for the unbalanced label class distribution (See Section~\ref{ssec:dataset}).

\begin{table}[ht]

\small
\setlength{\tabcolsep}{4pt} %
\renewcommand{\arraystretch}{1.4} %

\begin{tabular}{lcrrcrrcrrcrr}
\toprule
\multirow{4}{*}{\textbf{Dataset}} & \multicolumn{6}{c}{\textbf{ACL Anthology}} & \multicolumn{6}{c}{\textbf{CORD-19}} \\
\cmidrule(lr){2-7}
\cmidrule(lr){8-13}
{} & \multicolumn{3}{c}{\textbf{macro avg}} & \multicolumn{3}{c}{\textbf{micro avg}} & \multicolumn{3}{c}{\textbf{macro avg}} & \multicolumn{3}{c}{\textbf{micro avg}} \\ 
\cmidrule(lr){2-4}
\cmidrule(lr){5-7}
\cmidrule(lr){8-10}
\cmidrule(lr){11-13}
{} & \textbf{F1(std)} & \textbf{P} & \textbf{R} & \textbf{F1(std)}  & \textbf{P} & \textbf{R} & \textbf{F1(std)} & \textbf{P} & \textbf{R} & \textbf{F1(std)} & \textbf{P} & \textbf{R} \\

\midrule
\rowcolor{Gray} 

LSTM$_{baseline}$ &          .063  $\pm$.001 &      .069 &   .058 &      .290  $\pm$.004 &      \textbf{.761} &   .179 &      .128  $\pm$.001 &      .137 &   .121 &      .579  $\pm$.005 &      .758 &   .469
 \\

BERT                &          .256  $\pm$.002 &      .317 &   .238 &      .641  $\pm$.002 &      .719 &   .578 &      .387  $\pm$.011 &      \textbf{.619} &   .357 &      .822  $\pm$.002 &      .840 &   .806
 \\
\rowcolor{Gray} 

Covid-BERT               &          .270  $\pm$.006 &      .404 &   .253 &      .648  $\pm$.005 &      .715 &   .592 &      .394  $\pm$.010 &      .578 &   .364 &      .818  $\pm$.001 &      .836 &   .802
 \\

SciBERT       &         \textbf{.326  $\pm$.005 }&      \textbf{.458} &   \textbf{.303} &      \textbf{.678  $\pm$.002} &      .725 &   \textbf{.637} &      \textbf{.439  $\pm$.010} &      .560 &  \textbf{.401} &      \textbf{.833  $\pm$.003} &      \textbf{.846} &   \textbf{.820}
 \\
\rowcolor{Gray} 

RoBERTa                   &          .250  $\pm$.003 &      .285 &   .232 &      .626  $\pm$.003 &      .703 &   .564 &      .332  $\pm$.008 &      .473 &   .316 &      .820  $\pm$.001 &      .840 &   .801
 \\

XLNet               &          .263  $\pm$.011 &      .372 &   .250 &      .645  $\pm$.011 &      .705 &   .595 &      .362  $\pm$.025 &      .523 &   .345 &      .817  $\pm$.002 &      .832 &   .804
 \\\rowcolor{Gray} 
 
 ELECTRA     &          .245  $\pm$.005 &      .287 &   .228 &      .616  $\pm$.021 &      .693 &   .554 &      .280  $\pm$.001 &      .306 &   .276 &      .820  $\pm$.002 &      .840 &   .801
 \\ \bottomrule

\end{tabular}

\caption{Overall F1-score (with standard deviation), precision, and recall for macro and micro average of \methodsCountText methods for \acl and \cord. SciBERT yields best results in both datasets.}\label{tab:overall_results}
\end{table}

Given the overall scores, SciBERT is the best method with 0.326 macro-F1 and 0.678 micro-F1 on \acl, and with 0.439 macro-F1 and 0.833 micro-F1 on \cord.
All Transformer models outperform, in all metrics, the $\text{LSTM}_{baseline}$ except for the micro-precision on \acl.
The gap between macro and micro average results is due to discrepancies between the label classes (See Section~\ref{ssec:label_eval}).
BERT, SciBERT, and Covid-BERT perform better, on average, for \acl and \cord when compared to the baseline and the other Transformer-based models.
For \acl, the methods are ranked equal for both macro and micro. 
SciBERT presents the highest scores with a large margin, followed by Covid-BERT, XLNet, and BERT.
The lower performers are RoBERTa (0.626 micro-F1) and ELECTRA (0.616 micro-F1).
In terms of macro average, the methods present the same ranking for \cord and \acl except for BERT which outperforms XLNet.
Only for micro average on \cord the outcome is different, i.e., ELECTRA and RoBERTa achieve higher F1-scores than Covid-BERT and XLNet.
Even though Covid-BERT is fine tuned on \cord its performance yields a 0.818 micro-F1.

\subsection{Label Classes Evaluation} \label{ssec:label_eval}
We divide both datasets, \acl and \cord, into 11 label classes between positive and negative examples (Section~\ref{ssec:preprocessing} and~\ref{ssec:negative}).
Each class represents a different section in that a paper gets cited.
The section indicates in what aspects two papers are similar.
The aspects can also be ambiguous making the labels classification a hard task.
The following section investigates the classification performance with respect to the different label classes.
Table~\ref{tab:label_results} presents F1-score, precision, and recall of SciBERT for all 11 labels. Additionally, we include the overall results for single and multi-label samples (i.e., 2, and $\geq3$). %
The remaining methods from Table~\ref{tab:overall_results} present lower but proportional similar scores\footnote{The detailed data on the remaining methods is available together with the trained models in our GitHub repository.}. 

\begin{table}[ht]

\small
\setlength{\tabcolsep}{5pt} %
\renewcommand{\arraystretch}{1.2} %

\begin{tabular}{lrcrrlrcrr}
\toprule

\multicolumn{5}{c}{\textbf{ACL Anthology}} & \multicolumn{5}{c}{\textbf{CORD-19}} \\ 
\cmidrule(lr){1-5}
\cmidrule(lr){6-10}

      \textbf{Label} &   \textbf{Samples} &  \textbf{F1 (Std)} &  \textbf{P} &  \textbf{R} &     
      \textbf{Label} &   \textbf{Samples} &  \textbf{F1 (Std)} &  \textbf{P} &  \textbf{R} \\

\midrule
\rowcolor{Gray} 

   Background &    341 &     0.436 $\pm$ 0.045 &      0.651 &   0.329 &    Background &    113 &     0.617 $\pm$    0.042 &      0.655 &   0.588
 \\

   Conclusion &    289 &     0.000 $\pm$ 0.000 &      0.000 &   0.000 &    Conclusion &    250 &     0.274 $\pm$ 0.039 &      0.563 &   0.182
 \\
\rowcolor{Gray} 

   Discussion &    283 &     0.000 $\pm$  0.000 &      0.000 &   0.000 &    Discussion &   3314 &     0.636 $\pm$ 0.008 &      0.641 &   0.631
 \\

   Evaluation &    242 &     0.008 $\pm$  0.007 &      0.396 &   0.004 &   Future work &     42 &     0.032 $\pm$   0.064 &      0.150 &   0.018
 \\
\rowcolor{Gray} 

   Experiment &   1006 &     0.360 $\pm$   0.008 &      0.491 &   0.284 &  Introduction &   3777 &     0.644 $\pm$   0.004 &      0.669 &   0.620
 \\

 Introduction &   4069 &     0.527 $\pm$   0.005 &      0.576 &   0.486 &     Materials &    105 &     0.241 $\pm$    0.038 &      0.552 &   0.157
 \\
\rowcolor{Gray} 

       Methods &    179 &     0.014 $\pm$  0.028 &      0.208 &   0.007 &       Methods &    130 &     0.205 $\pm$   0.030 &      0.519 &   0.130
 \\

 Related work &   3150 &     0.638 $\pm$    0.012 &      0.660 &   0.617 &       Results &    227 &     0.322 $\pm$    0.021 &      0.558 &   0.227
 \\
\rowcolor{Gray} 

      Results &    295 &     0.015 $\pm$    0.011 &      0.475 &   0.008 &         Virus &     54 &     0.000 $\pm$      0.000 &      0.000 &   0.000
 \\

        Other &   5562 &     0.645 $\pm$     0.005 &      0.646 &   0.645 &         Other &  10788 &     0.876 $\pm$      0.002 &      0.872 &   0.879
 \\
\rowcolor{Gray} 

         \textit{None} &   6068 &     0.942 $\pm$      0.002 &      0.934 &   0.951 &          \textit{None} &   8270 &     0.979 $\pm$       0.001 &      0.980 &   0.977

\\

\cmidrule(lr){1-5}
\cmidrule(lr){6-10}

1 label &   15652 &    0.721 $\pm$      0.002 &      0.717 &   0.726 &          
1 label  &   22885 &     0.860 $\pm$      0.003 &      0.844 &   0.876 \\
\rowcolor{Gray} 
2 labels &   1968 &    0.540 $\pm$      0.003 	 &       	0.738 &   0.425 &         
2 labels &   1632 &     0.656 $\pm$      0.004 &      0.849 &   0.535 \\

$\geq3$ labels &   585 &    0.492 $\pm$     0.015 &     0.857 &   0.345 &         
$\geq3$ labels &   295 &     0.590 $\pm$      0.010 &      0.925 &   0.433 	 \\
\bottomrule

\end{tabular}

\caption{Results of SciBERT on \acl and \cord  datasets per label class, number of samples available (test set), F1-score (with standard deviation), precision, and recall.}\label{tab:label_results}
\end{table}

The \textit{None} has the highest F1-score (0.942 for \acl, 0.980 for \cord) with a large margin.
\textit{Other} shows the second-best F1-score, which in a similar-dissimilar classification scenario can be interpreted as an opposite class to the \textit{None} label. 
The remaining positive labels yield lower scores but also a lower number of samples.
Since we conduct a 4-fold cross validation the ratio of train and test samples is 75/25.
In \cord, 10,788 \textit{Other} test samples exist compared to 3,777 \textit{Introduction} samples, which is the most common section title (Table~\ref{tab:dataset_labels}). 
Still, the lower number of samples does not necessarily correlates with low accuracy. 
In \acl, the label \textit{Related work} (3,150 samples) yields higher scores when compared to \textit{Introduction} (4,069 samples) with a F1-score of 0.638 and 0.527 respectively. 
The label \textit{Background} in \cord has a F1-score of 0.617 despite having only 113 samples.
The results in Table~\ref{tab:label_results} show an impact from the label classes on the overall performance.
Six labels (\acl~-~\textit{Conclusion}, \textit{Discussion}, \textit{Evaluation}, and \textit{Methods}; \cord~-~\textit{Future work} and \textit{Virus}) have F1-scores between zero and 0.05.  
The discrepancy in the number of samples and difficulty in uncovering latent information from aspects contribute for the decrease in some labels' accuracy.
Even for domain experts, the location of whether one paper cites another, e.g., in \textit{Introduction} or \textit{Experiment}, is not trivial to predict.

The bottom rows in Table~\ref{tab:label_results} illustrate the effect of multi-labels. 
F1-scores decrease on both datasets as the number of labels increases. 
This is due to decreasing recall. 
The precision increases with more labels. %
Table~\ref{tab:multilabel} shows a portion of the distribution of multi-label samples in \cord and corresponding SciBERT predictions (the list is limited due to space restrictions). %
When two or more labels are present, SciBERT often correctly predicts one of the labels but not the others. 
For example, the two-label of \textit{Discussion} and \textit{Introduction} (D,I) has only 22\% test samples correct.
Still, SciBERT correctly predicts for the remaining samples one of the two labels, i.e., either \textit{Discussion} (35\%) or \textit{Introduction} (31\%).
We see comparable results for other multi-labels such as \textit{Discussion}, \textit{Introduction}, and \textit{Other} (D,I,O). 

\begin{table}[ht]
\renewcommand{\arraystretch}{1.3} %

\resizebox{0.95\textwidth}{!}{
\begin{tabular}{lrrrrrrrrrrrrrrrr}
\toprule
\multicolumn{2}{c}{\textbf{Ground Truth}} & \multicolumn{15}{c}{\textbf{Predictions}} \\
\cmidrule(lr){1-2}
\cmidrule(lr){3-17}
\textbf{Sections} & \textbf{Sample}
&  \textbf{\textit{N}} 
&   \textbf{B} 
&   \textbf{C}
&   \textbf{D} 
&   \textbf{I}
&    \textbf{O} &   \textbf{R} &  \textbf{C,O} &  \textbf{D,I} &  \textbf{D,O} &  \textbf{D,R} & \textbf{I,O} &  \textbf{O,R} &  \textbf{D,I,O} &  \textbf{D,O,R} \\
\midrule

\rowcolor{Gray} 
C,D   &    21 &   - & - & - &    1 &    6 &    7 & - &  - &    1 &  - &  - &    1 &  - &    - &    - \\

C,O   &    79 &   - & - &   2 &    1 &    2 &   58 & - &   13 &  - &  - &  - &    3 &  - &    - &    - \\

\rowcolor{Gray} 
\textbf{D,I}   &   \textbf{459} &     \textbf{1} & - & - &  \textbf{163} &  \textbf{146} &   \textbf{17} 
& - &  - &  \textbf{103} &    \textbf{7} &    \textbf{2} &    \textbf{9} &  - &     \textbf{10} &    - \\

D,O   &   351 &     1 &   2 & - &  102 &   30 &  120 &   1 &  - &   15 &   59 &    1 &    4 &    1 &      4 &    - \\

\rowcolor{Gray} 
D,R   &    65 &     1 & - & - &    6 &   10 &   10 & - &  - &    1 &    3 &   28 &  - &  - &    - &      1 \\

I,O   &   453 &     2 &   1 & - &   15 &  114 &  215 &   1 &  - &   12 &   16 &    1 &   62 &  - &      9 &    - \\

\rowcolor{Gray} 
D,I,O &   142 &     1 &   1 & - &   28 &   31 &   11 & - &  - &   33 &    8 &  - &   12 &  - &     14 &    - \\

D,O,R &    23 &   - & - & - &    5 &  - &    7 & - &  - &  - &    5 &    2 &  - &    1 &    - &      1 \\
\bottomrule
\end{tabular}}

\caption{\label{tab:multilabel} Confusion matrix of selected multi-labels for SciBERT on \cord (N=None, C=Conclusion, O=Other, D=Discussion, I=Introduction, R=Results). For example \textbf{(in bold)}, 459 test samples are assigned to \textit{Discussion} and \textit{Introduction} (D,I), of which 103 are correctly classified. The remaining samples are mostly classified as single-label, i.e., either \textit{Discussion} (163) or \textit{Introduction} (146).}
\end{table}

\subsection{Qualitative Evaluation}

To validate our quantitative findings, we qualitatively evaluate the prediction from SciBERT on \acl.
For each example in Table~\ref{tab:examples}, SciBERT predicts whether the seed cites the target paper and in which the section the citation should occur. 
We manually examine the predictions on their correctness.

\begin{table}[ht]
\renewcommand{\arraystretch}{1.5} %
\small
\begin{tabularx}{\textwidth}{p{0.05cm}XXp{1.7cm}p{1.8cm}}
\toprule
\textbf{ } & \textbf{Seed Paper} & \textbf{Target Paper}  & \textbf{Citation}                  & \textbf{Prediction}           \\ \midrule

\rowcolor{Gray} 
1 
& UKP: Computing Semantic Textual Similarity by Combining Multiple Content Similarity Measures \cite{Bar2012}
& SemEval-2012 Task 6: A Pilot on Semantic Textual Similarity \cite{Agirre2012}
& Other
& Introduction$\times$ 
\\ \hline

2 
& Query segmentation based on eigenspace similarity \cite{Zhang2009}
& Unsupervised query segmentation using generative language models and wikipedia \cite{Tan2008}
& Introduction, Experiment
& Introduction\checkmark, Experiment\checkmark
\\ \hline

\rowcolor{Gray} 
3 
& Transition-Based Parsing of the Chinese Treebank using a Global Discriminative Model \cite{ZhangYue2009}
& Enhancing Statistical Machine Translation with Character Alignment \cite{Xi2012}
& None
& Experiment$\times$
\\ \hline

4
& Experiments in evaluating interactive spoken language systems \cite{Polifroni1992}
& Evaluating information presentation strategies for spoken recommendations \cite{Winterboer2007}
& None
& Introduction$\times$, Other$\times$
\\ \hline

\rowcolor{Gray} 
5
& Similarity-based Word Sense Disambiguation \cite{Karov1998}
& Targeted disambiguation of ad-hoc, homogeneous sets of named entities \cite{Wang2012}
& None
& None\checkmark
\\ \hline

6
& SciSumm: A Multi-Document Summarization System for Scientific Articles \cite{Agarwal2011}
& Improving question-answering with linking dialogues \cite{Gandhe2006}
& None
& None\checkmark
\\ \bottomrule

\end{tabularx}

\caption{\label{tab:examples} Example labels of research paper pairs (seed and target) as defined by citations and as predicted using SciBERT. Based on the test set, correct predictions are marked with~\checkmark, invalid ones with $\times$.}
\end{table}

The first example of \newcite{Bar2012} and \newcite{Agirre2012} is a correct prediction.
Given the ground truth, the aspect is \textit{Other} (the citation occurs in a section called ``Results on Test Data"). %
We assess \textit{Introduction} as a potential valid prediction since 
\newcite{Bar2012} is a submission to the shared-task described in \newcite{Agirre2012}.
Therefore, one could have cited it in the introduction. 
All predictions in example~2 are correct. 
Compared to the other examples, we consider example~2 a simple case as both papers mention their topic (i.e., query segmentation) in the title and in the first sentence of the abstract (hint for \textit{Introduction} label).
Both abstracts of example~2 also refer to ``mutual information and EM optimization'' as their methods.
In example~3, \newcite{ZhangYue2009} and \newcite{Xi2012} do not share any citation.
Hence, the paper pair is assigned with the \textit{None} label according to the ground truth data even though they are topically related.
\newcite{ZhangYue2009} and \newcite{Xi2012} are both about Chinese machine translation.
Still, we disagree with our model's prediction of \textit{Experiment} since the two papers conduct different experiments making \textit{Experiment} an invalid prediction.
Example~4's predictions are correct. 
\newcite{Polifroni1992} is published before \newcite{Winterboer2007} and, therefore, a citation cannot exist. 
Nonetheless, the two papers cover a related topic. 
Thus, one could expect a citation of \newcite{Polifroni1992} in \newcite{Winterboer2007} in the introduction section as SciBERT predicted.
Our model finds this semantic similarity given their latent information on the topic.
Example~5-6 present two pairs for which \textit{None} was correctly predicted according to the ground truth.
\newcite{Agarwal2011} and \newcite{Gandhe2006} from Example~6 are topically unrelated as their titles already suggest.
However, \newcite{Karov1998} and \newcite{Wang2012} on Example~5 share the topic of \emph{disambiguation}. 
Thus, we would agree with the prediction of a positive label.

In summary, the qualitative evaluation does not contradicts our quantitative findings.
SciBERT distinguishes documents at a higher level and classifies which aspects makes them similar.
In addition to traditional document similarity, the aspect-based predictions allow to asses how two papers relate to each other at a semantic level. 
For instance, whether two papers are similar in the aspects of \textit{Introduction} or \textit{Experiment} is a valuable information, especially in literature reviews.

\section{Discussion}
\label{sec:discussion}

In our experiments, SciBERT outperforms all other methods in the pairwise document classification.
We observe in-domain pretraining and NSP objective often lead to higher F1-scores.
Transferring generic language models to a specific domain usually decreases the performance in our experiments.
A possible explanation for this is the narrowly defined vocabulary in \acl or \cord.
\newcite{Beltagy2019} and \newcite{Lee2019} have also explored the transfer learning between domains with similar findings.
Covid-BERT seems to be an exception as it yields lower results (micro-F1) than BERT on \cord even though Covid-BERT was fine-tuned on \cord.
We observe the language model fine-tuning in Covid-BERT does not guarantee a higher performance compared to pretraining from scratch in SciBERT.
However, Covid-BERT's authors provide too little information to give a proper explanation for its performance.
Apart from in-domain pretraining, the NSP objective has a positive effect on the models.
All BERT-based systems, which use NSP, outperform the models that excluded NSP (XLNet, RoBERTa, and ELECTRA). 
We attribute the positive effect of NSP to its similarity to our task since both are sequence pair classification tasks.
Table~\ref{tab:overall_results} and \ref{tab:label_results} show variance among labels and both datasets.
The larger number of training samples in \cord (36\%) may have contributed to a higher performance in comparison to \acl.
An unbalanced class distribution and different challenges of the labels cause the performance to differ between the label classes.
The high F1-scores of above 0.9 for negative samples are expected since the \textit{None} label is essentially an aspect-free similarity or citation prediction problem.
Transformer models have been shown to perform well in these two problems~\cite{Reimers2019,Cohan2020}.
Besides the unbalanced distribution of training samples, we attribute the differences among positive labels to their ambiguity and to the different challenges posed by the label classes.
Authors often diverge when naming their section titles (e.g., \textit{Results}, \textit{Evaluation}), thus, increasing the challenge of labeling the different aspects of a paper.
This also contributes to the high number of \textit{Other} samples.
Some sections are also content-wise more unique than others.
An \textit{Introduction} section usually contains different content than a \textit{Results} section.
The content difference makes some sections and the corresponding label classes easier to distinguish and predict than others.
We suspect the poor performance for \textit{Future work} is due to little or no information about them in the titles or abstracts.

Our main research objective in this paper is to explore methods that are capable to incorporate aspect information into the traditional similar-dissimilar classification.
In this regard, we deem the results as promising.
In particular, the  micro-F1 score of 0.86 of SciBERT for the \cord dataset is encouraging. 
Our qualitative evaluation indicates that SciBERT's predictions can correctly identify similar aspects of two research papers.
In order to verify if our first indication generalizes, a large qualitative survey needs to be conducted.
Furthermore, we observe that label classes with little training data performed poorly.
For example, \textit{Conclusion} and \textit{Discussion} have a zero F1-score for \acl whereas for the larger \cord dataset \textit{Discussion} yields 0.636 F1.
We anticipate that more training data leads to more correct predictions.

\section{Conclusion}
\label{sec:conclusions}
In this paper, we apply pairwise multi-label multi-class document classification on scientific papers to compute a aspect-based document similarity score. 
We treat section titles as aspects of paper and label citations occurring in these sections accordingly. 
The investigated models are trained to predict citations and the respected label based on the paper's title and abstract. %
We evaluate the Transformer models BERT, Covid-BERT, SciBERT, ELECTRA, RoBERTa, and XLNet and a LSTM baseline over two scientific corpora, i.e., \acl and \cord. 
Overall, SciBERT performed best in our experiments.
Despite the challenging task, SciBERT predicted the aspect-based document similarity with F1-scores of up to $0.83$.
SciBERT's performance motivates further research in this direction. 
It seems reasonable to include the aspect-based document similarity task as a new pretraining objective in the Transformers architecture.
This new objective could be integrated in similar fashion as the binary citation prediction objective \newcite{Cohan2020} proposed.
As future work, we plan to integrate the aspect-based document similarity into a recommender system. 
Thus, enabling a large user study to confirm our first indications that aspect-based document similarity indeed helps users to find more relevant recommendations.
However, our extensive empirical analysis already demonstrates that Transformers are well-suited to correctly compute the aspect-based document similarity for research papers.

\section*{Acknowledgements}

We  would  like  to  thank  all  reviewers and Christoph Alt for their comments and valuable feedback.
The research presented in this article is funded by the German Federal Ministry of Education and Research (BMBF) through the project QURATOR (Unternehmen Region, Wachstumskern, no.~03WKDA1A).

\bibliographystyle{coling}
\bibliography{paper}

\end{document}